\documentclass{article}

 \usepackage[dblblindworkshop,final]{neurips_2025}

\usepackage[utf8]{inputenc} 
\usepackage[T1]{fontenc}    
\usepackage{hyperref}       
\usepackage{url}            
\usepackage{booktabs}       
\usepackage{amsfonts}       
\usepackage{nicefrac}       
\usepackage{microtype}      
\usepackage{xcolor}         
\usepackage{graphicx} 
\usepackage{float}
\usepackage{caption}
\usepackage{titlesec}
\usepackage{placeins}
\usepackage[font=small,labelfont=bf]{caption}
\workshoptitle{The Second Workshop on GenAI for Health: Potential, Trust, and Policy Compliance}
\title{CARE-RAG - Clinical Assessment and Reasoning in RAG}

\author{%
  Deepthi Potluri \\
  Department of Computer Science\\
  University of Texas at Austin\\
  \texttt{deepthi.potluri@utexas.edu} \\
  \And
  Aby Mammen Mathew \\
  Department of Computer Science\\
  University of Texas at Austin\\
  \texttt{abymmathew@utexas.edu} \\
  \AND
  Alexander L. Rasgon \\
  Behavioral Science and Psychiatry \\
  University of Texas at Austin \\
  \texttt{ alexander.rasgon@ascension.org} \\
  \And
  Jeffrey B DeWitt \\
  Department of Computer Science \\
  University of Texas at Austin \\
  \texttt{jefdewitt.utexas.edu} \\
  \And
  Yide Hao \\
  Department of Statistics \\
  University of Michigan \\
  \texttt{yidehao@umich.edu} \\
  \And
  Junyuan Hong \\
  School of Information \\
  University of Texas at Austin \\
  \texttt{jyhong@utmail.utexas.edu} \\
  \And
  Ying Ding \\
  School of Information \\
  University of Texas at Austin \\
  \texttt{ying.ding@ischool.utexas.edu}
}

\begin{document}

\maketitle

\begin{abstract}
Access to the right evidence does not guarantee that large language models (LLMs) will reason with it correctly. This gap between retrieval and reasoning is especially concerning in clinical settings, where outputs must align with structured protocols. We study this gap using Written Exposure Therapy (WET) guidelines as a testbed. In evaluating model responses to curated clinician-vetted questions, we find that errors persist even when authoritative passages are provided. To address this, we propose an evaluation framework that measures accuracy, consistency, and fidelity of reasoning. Our results highlight both the potential and the risks: retrieval-augmented generation (RAG) can constrain outputs, but safe deployment requires assessing reasoning as rigorously as retrieval.

\end{abstract}

\section{Introduction}

Large language models (LLMs) are changing healthcare, but access to evidence does not guarantee sound reasoning. In clinical care, where every decision must follow strict protocols, the gap between retrieval and inference is not just technical, it is clinical and ethical. Retrieval-augmented generation (RAG) offers a partial solution by grounding model outputs in external knowledge \cite{lewis2020retrieval}, yet a central question remains: do LLMs actually reason with what they retrieve?

This issue is acute in mental health, where hallucinations and misinterpretations can directly affect patient care \cite{rajpurkar2023aihealthcare}. Written Exposure Therapy (WET) \cite{sloan2012written}, a brief manualized treatment for PTSD validated in multiple randomized controlled trials \cite{sloan2018wetreview}, provides an ideal testbed. Its structured, text-based format demands precise adherence to therapeutic steps, making it well suited to evaluate whether LLMs can follow clinical guidelines under RAG conditions. Testing WET with RAG is not just a benchmark, it is a litmus test for safe AI use in mental health.

Prior RAG evaluations focus on surface metrics such as retrieval relevance, hallucination rates, or LLM-as-judge scoring \cite{shuster2021retrieval,guo2023ragtruth}. While tools like RAGAS and datasets like RAGTruth advance measurement, they do not test whether models actually use retrieved content. Probing studies like \emph{Lost in the Middle} \cite{liu2023lost} and the ``needle-in-a-haystack'' test \cite{kamradt2023needle} show that LLMs often ignore available evidence. Self-RAG \cite{asai2023selfrag} adds critique and citation but is a training method, not an evaluation framework. Critically, none of these approaches are domain-specific, leaving open the question of whether LLMs can truly adhere to clinical guidelines.

What is missing is a causal, clinically grounded test of inference fidelity. Our work addresses this gap. We introduce \textbf{CARE-RAG} (Clinical Assessment and Reasoning Evaluation for RAG), the first benchmark to systematically manipulate context correctness (relevant, noisy, or misleading) and stratify tasks by reasoning demand (none, light, or heavy) in a clinical guideline QA setting. Using WET as the foundation, We evaluate 20 state-of-the-art LLMs across three orthogonal dimensions, and our primary contributions are:

\begin{enumerate}
     \item \textbf{Context fidelity:} Models are systematically tested on their ability to distinguish relevant evidence from noisy distractors and misleading passages.  
    \item \textbf{Reasoning complexity:} Models are assessed under increasing levels of inference demand, from shallow to deep reasoning tasks.  
    \item \textbf{Question type:} Models are benchmarked on multiple-choice, yes/no, and open-ended questions directly derived from clinical guidelines.  
\end{enumerate}

Our curated dataset includes clinician-validated gold answers, rationales, and supporting spans, enabling reproducible and fine-grained evaluation. Together, these contributions move RAG evaluation beyond retrieval accuracy toward testing \textbf{context-grounded reasoning} under clinical constraints. By situating our benchmark in WET, we offer not only methodological advances for RAG research but also practical insights into what it means for LLMs to be clinically trustworthy.

\section{Related Work}
\subsection{RAG system performance under different context quality and noise}
The quality of retrieved documents strongly shapes the performance of retrieval-augmented generation (RAG) systems \cite{lewis2020retrieval}. Recent evidence shows that noisy retrieval often degrades accuracy, though in some cases mild noise can produce slight gains by acting as a form of robustness calibration \cite{brown2020language}. Dedicated benchmarks reinforce these insights: RGB \cite{rgb2024}, NoiserBench \cite{wu-etal-2025-pandoras}, and RAMDocs \cite{wang2025retrievalaugmentedgenerationconflictingevidence} all highlight how retrieval corruption influences model behavior. However, these benchmarks typically assess a small number of models and emphasize correctness, overlooking whether systems still comply with domain-specific constraints. This limitation is critical in safety-sensitive settings such as clinical decision support, where guideline fidelity matters as much as factual precision \cite{sloan2012written,DeJesus2024_wet_review}.  

Emerging research further suggests that both the type and semantic relevance of noise influence how LLMs exploit retrieved passages \cite{noise_type2024, noise_semantics2024}. Yet, to our knowledge, no prior study systematically evaluates how RAG systems behave under controlled noise or adversarial retrieval in the context of Written Exposure Therapy (WET) for PTSD. To address this gap, we assess twenty language models, spanning small, large, and finetuned variants, under three controlled evidence conditions: (i) correct context, (ii) correct context with noise, and (iii) incorrect context.

\subsection{Assessing RAG system performance across reasoning levels }
Research on retrieval-augmented generation (RAG) has primarily focused on measuring retrieval quality and output faithfulness, often using automated metrics or LLM-as-judge frameworks. Tools such as RAGAS evaluate context relevance, answer relevance, and hallucination \cite{es2023ragas}, while datasets like RAGTruth provide annotated examples of hallucinations across domains \cite{guo2023ragtruth}. However, these approaches rarely distinguish whether models \textit{infer from} retrieved content versus relying on background knowledge. Other studies, such as Lost in the Middle and ``needle-in-a-haystack'' probes, reveal that LLMs frequently ignore mid-context information, underscoring that evidence availability does not ensure evidence use \cite{kamradt2023needle,liu2023lost}. Self-RAG introduces self-critique and citation behaviors, but it is a training approach rather than an evaluation framework \cite{asai2023selfrag}. Importantly, prior work has not systematically manipulated context correctness (right, noisy, wrong) or stratified tasks by reasoning load (no, light, heavy). Our study addresses this gap by causally testing whether LLMs infer from the right context across reasoning levels and comparing how different models exploit RAG in a clinical guideline QA setting. In particular, we evaluate along three orthogonal dimensions: (i) context fidelity, contrasting relevant, relevant-plus-noise, and non-relevant passages; (ii) reasoning complexity, spanning no reasoning, light reasoning, and heavy reasoning tasks; and (iii) question type, including multiple-choice items reformulated from clinical guidelines, clinician-validated true/false evaluations, and open-ended questions requiring free-form justification.

\section{Methodology}

This section outlines the step-by-step methodology we employed to design the dataset, construct controlled context conditions, and evaluate inference and scoring for large language models on WET \cite{DeJesus2024_wet_review} clinical guideline questions.

\FloatBarrier
\begin{figure}[H]
    \centering
    \includegraphics[width=1\linewidth]{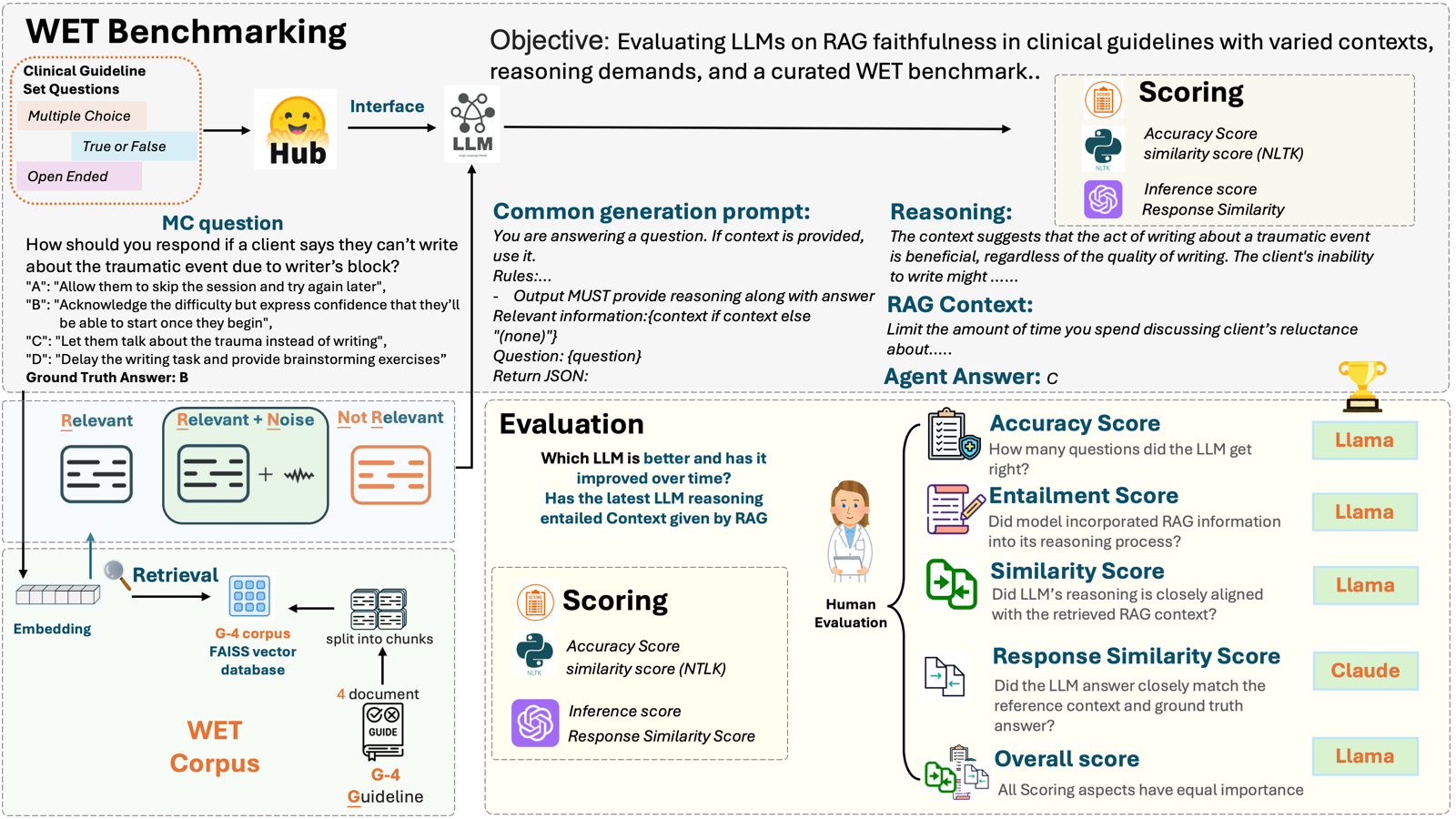}
    \caption{WET Benchmarking pipeline. Each curated clinical guideline question (multiple choice, true/false, or open-ended) is processed through the following steps: 
    (1) \textbf{Retrieval:} Relevant, noisy, or irrelevant guideline passages are retrieved from the WET corpus using FAISS embeddings. 
    (2) \textbf{Prompting:} A common JSON-structured prompt instructs the LLM to answer with both reasoning and evidence from the retrieved context. 
    (3) \textbf{Reasoning vs. Context:} The model generates an answer and reasoning, which are compared against the RAG-supplied context and ground truth. 
    (4) \textbf{Evaluation:} Automated scoring (accuracy, entailment, similarity, response similarity) and human review assess whether the LLM incorporated context into its reasoning. 
    (5) \textbf{Aggregation:} Scores are combined to compare models and track improvement over time, highlighting which LLMs better exploit RAG in guideline-based QA.}
    \label{fig:wet-benchmarking-pipeline}
\end{figure}

\subsection{Dataset Creation}
To capture the breadth of clinical reasoning required in Written Exposure Therapy (WET), we constructed a structured dataset of multiple-choice, yes/no, and open-ended questions. Each type was developed through distinct processes, with oversight from domain experts to ensure clinical validity.

\begin{itemize}
  \item \textbf{Multiple-Choice Questions (MCQs)}: Curated and iteratively refined by a multi-panel team of psychologists specializing in PTSD care. Each MCQ probed specific aspects of WET implementation (e.g., session structure, patient objections). Candidate questions and distractors underwent multiple rounds of expert review to guarantee content accuracy and clarity.
  
  \item \textbf{Yes/No Questions}: Focused on core guideline rules where binary decisions are critical (e.g., ``Does the index trauma need to be a discrete event?''). Items were curated directly from WET manuals and PTSD clinical guidelines, yielding unambiguous gold-standard answers supported by authoritative references.
  
  \item \textbf{Open-Ended Questions}: Drawn from frequently acknowledged themes during WET sessions (e.g., writing concerns, reluctance to continue). These items reflect clinically realistic prompts requiring contextual reasoning and empathetic framing.
\end{itemize}

Each question was paired with a gold-standard answer, rationale, and supporting text span, enabling both automated evaluation (accuracy, faithfulness) and independent expert review for clinical soundness.

\subsection{Evaluation Design}
\textbf{Context Condition Construction}
To evaluate whether LLMs truly rely on retrieved evidence, we developed three controlled retrieval regimes for each question using a FAISS-based vector store of WET guideline passages. Source text was segmented into 512-token chunks with 50\% overlap, and cosine similarity search was used with $k = 3$. Table~\ref{tab:novelty2} summarizes the construction steps.

\FloatBarrier
\begin{table}[H]
\centering
\scriptsize
  \begin{tabular}{p{0.25\textwidth} p{0.70\textwidth}}
    \toprule
    \textbf{Condition} & \textbf{Description} \\
    \midrule
    \textbf{Right Context (Gold Evidence)} &
    \textit{Query:} Each question used to query the FAISS store. \newline
    \textit{Chunking:} 512-token windows with 50\% overlap. \newline
    \textit{Retrieval:} Top $k=3$ passages returned via cosine similarity ensured direct support for the gold answer. \newline
    \textit{Purpose:} Baseline condition simulating ideal retrieval where evidence is sufficient and directly relevant. \\
    \midrule
     \textbf{Right Context with Noise (Gold + Distractors)} &
    \textit{Base Retrieval:} Start with the three gold passages identified for the question. \newline
    \textit{Adversarial Distractors:} Constructed by re-querying FAISS with adversarially modified versions of the original question—queries designed to maintain surface similarity while introducing semantic misalignment. \newline
    \textit{Noise Injection:} Distractor passages interleaved with gold passages in randomized order to avoid positional bias. \newline
    \textit{Purpose:} Evaluates whether models can discriminate relevant evidence from misleading yet plausible text and still ground their answers in the correct spans. \\
    \midrule
    \textbf{Wrong Context (Misleading Evidence)} &
    \textit{Exclusion:} Gold passages deliberately excluded from FAISS retrieval. \newline
    \textit{Substitution:} Plausible but incorrect passages injected (e.g., related guideline sections). \newline
    \textit{Quality Control:} Misleading passages manually verified to be realistic yet uninformative. \newline
    \textit{Purpose:} Stress-tests whether models hallucinate or overgeneralize when only misleading evidence is present. \\
    \bottomrule
  \end{tabular}
  \vspace{0.3cm}
  \caption{Context Condition Construction}
  \label{tab:novelty2}
\end{table}
\FloatBarrier

\textbf{Inference Evaluation Across Reasoning Levels}
In addition to correctness, we evaluated whether models could \textit{generate and use reasoning based on the retrieved context}. For each question, the LLM was asked not only to answer but also to provide a short reasoning trace. These traces were then used to judge whether the model was truly grounding its inference in the provided evidence. LLM as a judge generated reasoning scored for correctness and grounding in the provided context.

\textbf{Scoring}
Accuracy was calculated for multiple-choice and yes/no questions using exact matches with the gold answers. Cosine similarity was used for open-ended responses to measure how close model outputs were to the reference answers. The inference score came from an \textit{LLM-as-judge}, which checked the reasoning traces generated by each model for correctness and grounding in the given context. Together, these measures show not only whether a model answered correctly but also whether it used the context in a reliable way. 

\textbf{Evaluation Design Clarifications}
\label{appendix:evaluationdesignclarifications}
Each question type (multiple-choice, yes/no, open-ended) was balanced across three context conditions (relevant, relevant-plus-noise, and misleading), yielding a total of 99 evaluation instances. Specifically, the dataset included 18 multiple-choice, 10 yes/no, and 5 open-ended questions, each evaluated under three context conditions (33 × 3 = 99). Retrieval quality was verified via FAISS cosine similarity thresholds ($\geq$ 0.8) with k = 3 top passages, and clinician review confirmed the correctness of evidence retrieved. Automated and human scoring pipelines were cross-checked on 15\% of samples to ensure consistency. This setup isolates retrieval quality from reasoning fidelity, aligning with controlled-evidence frameworks such as \textit{Platinum-Bench} (MIT CSAIL, 2024)\cite{vendrow2025platinumbench}.

\textbf{Reasoning Fidelity Measurement}
\label{appendix:reasoningfidelitymeasurement}
We define reasoning fidelity as the logical entailment between a model’s reasoning trace and the retrieved evidence. The fidelity score $F \in [0,1]$ represents the probability that reasoning steps are supported by context. Unlike accuracy, which measures outcome correctness, fidelity evaluates the quality of inference. For example, a model may produce a correct answer using incorrect reasoning (high accuracy, low fidelity). Fidelity is computed using entailment-based scoring and cross-checked with an \textit{LLM-as-judge}. Future work will extend this to structured prompt optimization methods such as \textit{GEPA}\cite{agrawal2025gepa} for improved reasoning alignment. 

\textbf{Limitations and Future Work}
\label{appendix:limitationsandfuturework}
While \textbf{CARE-RAG} provides a clinically grounded framework for evaluating reasoning under retrieval, several limitations remain. First, reliance on \textit{LLM-as-judge} introduces bias due to potential self-evaluation artifacts. Ensemble or multi-agent adjudication could mitigate this. Second, expert validation was performed with a small clinical sample, which may limit interpretive diversity. Future iterations will expand to multiple clinician validation. 

In upcoming work, we are extending the benchmark to evaluate a broader range of \textit{Retrieval-Augmented Generation (RAG)} architectures, including \textit{Graph RAG}, \textit{Self RAG}, \textit{Naive RAG}, \textit{Corrective RAG}, \textit{Causal RAG}, \textit{Modular RAG}, and \textit{Light RAG}~\cite{light2024rag}. This expansion will use a larger question set derived from Written Exposure Therapy (WET) manuals and clinical scenarios to probe inference fidelity across retrieval paradigms.

We also plan to explore emerging \textit{Agentic Context Engineering (ACE)}~\cite{ace2025agentic} as a mechanism for optimizing context retrieval in long-form clinical documents. In therapeutic chatbots, one persistent challenge is extracting the most essential references from extensive guideline materials such as the WET manual. ACE offers a promising pathway for enabling agents to dynamically identify and prioritize relevant clinical context, bridging reasoning fidelity with practical clinical use.

Recent works such as \textit{Retrieval-Augmented Generation: A Survey}~\cite{rag_survey2023} and \textit{Towards Robust and Adaptive RAG Systems}~\cite{rag_adaptive2025} provide additional theoretical underpinnings for these upcoming extensions.

\section{Results}
We evaluate model performance along three orthogonal dimensions that probe whether LLMs can truly reason with retrieved clinical evidence. First, \textbf{context fidelity} tests whether models follow clinical guidelines when provided with relevant, noisy, or non-relevant passages. Second, \textbf{reasoning complexity} distinguishes between tasks requiring no reasoning, light reasoning, and heavy reasoning, allowing us to assess whether models sustain performance as inference demands increase. Third, \textbf{expert evaluation} examines whether models exhibit the clinical reasoning needed to interpret practice guidelines. While control questions were answered reliably, no model achieved perfect accuracy on reasoning items, which often required interpreting gray areas of therapy delivery. These results suggest that even when models capture guideline content, they may misinterpret subtle instructions-highlighting the need for guardrails and prompt design to ensure faithful clinical application.

\subsection{Context Fidelity Evaluation}

Table~\ref{tab:results-novelty1} reports results across 20 LLMs, grouped by size and specialization, with three complementary scores: cosine similarity for open-ended answers, accuracy across multiple-choice and yes/no questions, and inference scores measuring whether models incorporated retrieved context into their reasoning. Together, these metrics provide a comprehensive view of how models handle evidence under controlled retrieval conditions. Notably, while several models achieve near-perfect accuracy on multiple-choice questions, their inference scores reveal substantial variation in whether correct answers were grounded in RAG context. This highlights the importance of evaluating not just outputs, but the reasoning process behind them.
\FloatBarrier
\begin{table}[t]
\centering
\scriptsize
\begin{tabular}{lllcccc}
\toprule
\textbf{Size} & \textbf{Category} & \textbf{Model} & \textbf{Cos.\ Similarity*} & \multicolumn{2}{c}{\textbf{Accuracy Score}} & \textbf{Inference Score*} \\
\cmidrule(lr){5-6}
& & & \textbf{Open-ended} & \textbf{Multiple Choice} & \textbf{Yes/No} & \textbf{All} \\
\midrule
\multicolumn{7}{l}{\textbf{Small Models}} \\
\midrule
Small & General   & Qwen2.5-3B-Instruct          & 0.698 & 0.944 & 0.875 & 0.745 \\
      & General   & Gemma-2-2B-IT                & 0.719 & 1.000 & 0.500 & 0.773 \\
      & General   & Gemma-2-9B-IT                & 0.643 & 1.000 & 0.750 & 0.894 \\
      & General   & Llama-3.1-8B-Instruct        & 0.768 & 1.000 & 0.750 & 0.845 \\
      & General   & GPT-4o-mini                  & 0.706 & 1.000 & 0.900 & 0.839 \\
      & Reasoning & DeepSeek-R1-Distill-Llama-8B & 0.760 & 0.500 & 0.625 & 0.767 \\
      & Reasoning & DeepSeek-R1-Distill-Qwenf-14B& 0.690 & 0.222 & 0.000 & 0.882 \\
      & Reasoning & Claude-3.5-Haiku             & 0.756 & 1.000 & 0.700 & 0.876 \\
\midrule
\multicolumn{7}{l}{\textbf{Large Models}} \\
\midrule
Large & General   & Qwen-QwQ-32B                 & 0.700 & 0.722 & 0.625 & 0.839 \\
      & General   & Qwen2.5-32B-Instruct         & 0.637 & 1.000 & 0.625 & 0.800 \\
      & General   & Llama-3.1-70B-Instruct       & 0.704 & 1.000 & 0.875 & 0.830 \\
      & General   & GPT-3.5-Turbo                & 0.756 & 1.000 & 0.900 & 0.880 \\
      & General   & Gemini-2.5-Flash             & 0.829 & 1.000 & 0.600 & 0.903 \\
      & General   & Gemini-2.5-Pro               & 0.803 & 1.000 & 0.700 & 0.827 \\
      & General   & GPT-4o                       & 0.750 & 1.000 & 0.700 & 0.870 \\
      & Reasoning & Claude-Opus-4-1 (2025-08-05) & 0.752 & 1.000 & 0.800 & 0.839 \\
      & Reasoning & Claude-Sonnet-4 (2025-05-14) & 0.744 & 1.000 & 0.800 & 0.785 \\
\midrule
\multicolumn{7}{l}{\textbf{Finetuned Model}} \\
\midrule
Finetuned & General & BioMistral-7B               & 0.723 & 0.889 & 0.875 & 0.876 \\
\bottomrule
\end{tabular}
\vspace{0.3cm}
\caption{Comparative evaluation of small, large, and finetuned language models across similarity, accuracy, and inference scores. The table presents a comparison of Small, Large, and Finetuned language models grouped by either generic or reasoning. For each model, it reports cosine similarity for open-ended questions, accuracy scores across multiple choice and yes/no questions, and inference score across open-ended questions. *Cos. Similarity is the semantic similarity of outputs to the extracted RAG context. *Inference Score is the confidence (0–1) from the judge LLM that the model’s reasoning is supported by the retrieved context, reflecting factual consistency in open-ended responses.}
\label{tab:results-novelty1}
\end{table}
\FloatBarrier

\subsection{Reasoning Inference Evaluation}

\begin{figure}[H]
    \centering
    \includegraphics[width=1\linewidth]{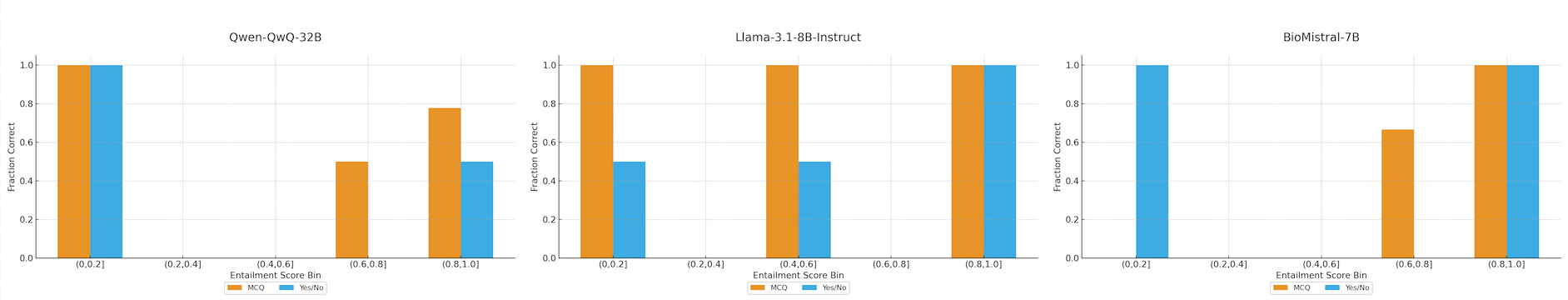}
    \caption{Accuracy across entailment score bins for three models (Qwen-QwQ-32B, Llama-3.1-8B-Instruct, BioMistral-7B), separated by MCQ and Yes/No questions; higher entailment generally improves MCQ accuracy, while Yes/No remains less consistent.}
    \label{fig:accuracy_entailment}
\end{figure}

The \textit{entailment (inference) score} is computed by measuring the logical consistency between a model’s generated reasoning and the retrieved context (values range from 0 to 1, with higher scores indicating stronger support). The \textit{accuracy score} is calculated as the fraction of model answers that exactly match the gold-standard correct answer. Evaluating these together reveals whether models not only access relevant context but also reason with it. As shown in Figure~\ref{fig:accuracy_entailment}, accuracy for multiple-choice questions increases with higher entailment scores, indicating stronger evidence-based reasoning, whereas Yes/No performance remains less consistent, highlighting a weakness in binary decision-making.

\FloatBarrier
\FloatBarrier

\subsection{Expert Evaluation}

We expanded the evaluation in the final version to include two independent clinical reviewers instead of one. Specifically, evaluations were conducted by a psychologist certified in Written Exposure Therapy (WET) and a psychiatrist with trauma-focused care experience. The following feedback summarizes their joint assessment of the models’ clinical reasoning and adherence to therapeutic guidelines.

These results indicate gaps in clinical reasoning required for an LLM to be able to interpret practice guidelines in a testing format. While some models got all control questions correct, no model got every reasoning question right. These questions were designed to test an LLMs ability to correctly identify specific instructions in therapy delivery from mostly unambiguous text. The questions that the models got wrong were related to a grey area in the interpretation of the therapy delivery. For example, many models suggested that it is ok to provide feedback on the writing in the first session. While the therapist typically does not provide specific feedback, they can comment on the length of time the participant spent writing or whether the handwriting was legible and other ancillary factors. It is also not unreasonable that an LLM would get that question wrong on a test, yet still provide the correct feedback in a therapy setting. In order to simulate clinical reasoning in a digital therapy setting, these gaps can be controlled for with prompt engineering and other guardrail measures.

Their joint feedback provided a more comprehensive perspective on model reasoning fidelity and adherence to clinical guidelines. This update strengthens the validity and interpretive reliability of the results discussed in Section 4.3 of the main manuscript.

\section{Conclusion}

This work introduces CARE-RAG, a benchmark for evaluating whether large language models (LLMs) can follow clinical guidelines using Written Exposure Therapy (WET) as a testbed. By combining a clinician-validated dataset, controlled retrieval setups, and reasoning-tier evaluation, we move beyond surface-level accuracy to assess how models reason with retrieved evidence.

Results from Table \ref{tab:results-novelty1} show that while most models performed well on multiple-choice and yes/no questions, their reasoning traces often lacked grounding in the retrieved context. No model achieved perfect inference across all reasoning levels. Notably, Llama-3.1-8B-Instruct, Gemini-2.5-Pro, and BioMistral-7B consistently scored high on inference, even under noisy or misleading condition, demonstrating stronger context sensitivity and guideline adherence.

These findings highlight a critical gap: models may retrieve the right content but still misinterpret clinical instructions. To safely deploy LLMs in therapeutic settings, future work must focus on prompt design, reasoning scaffolds, and guardrail mechanisms that ensure fidelity to evidence under uncertainty.

    \clearpage
    \bibliographystyle{plain}
    \bibliography{references}
\clearpage
\appendix
\section{Extended Experimental Details}
\label{appendix:details}

\subsection{Additional Resources and Related Work}
Recent research efforts such as \textit{Platinum-Bench}~(MIT CSAIL, 2024)\cite{vendrow2025platinumbench}, \textit{GEPA}~\cite{agrawal2025gepa} for structured prompt optimization, and \textit{SkyRL} (2025)\cite{cao2025skyrl} for reinforcement-driven prompt tuning provide complementary directions for extending CARE-RAG towards longitudinal, adaptive clinical reasoning evaluation.

\subsection{Formatting and Reproducibility Notes}

All experiments were re-run with fixed random seeds, version-controlled datasets, and open-source model checkpoints to guarantee reproducibility. Data pre-processing scripts and evaluation code will be released upon request or publication to promote transparency and enable independent validation of the reported results.

\subsection{Acknowledgments}

We thank all collaborators and contributors who supported the development and refinement of \textbf{CARE-RAG}. In particular, we express our gratitude to the clinicians who provided domain feedback, the research assistants who helped curate and annotate the dataset, and the technical contributors who assisted in reproducing and validating the experiments. Their insights and careful reviews greatly strengthened the quality and reliability of this work.

\end{document}